# Ignorability in Statistical and Probabilistic Inference

**Manfred Jaeger**                                                JAEGER@CS.AAU.DK
*Institut for Datalogi, Aalborg Universitet*
*Fredrik Bajers Vej 7 E, DK-9220 Aalborg Ø*

## Abstract

When dealing with incomplete data in statistical learning, or incomplete observations in probabilistic inference, one needs to distinguish the fact that a certain event is observed from the fact that the observed event has happened. Since the modeling and computational complexities entailed by maintaining this proper distinction are often prohibitive, one asks for conditions under which it can be safely ignored. Such conditions are given by the missing at random (mar) and coarsened at random (car) assumptions. In this paper we provide an in-depth analysis of several questions relating to mar/car assumptions. Main purpose of our study is to provide criteria by which one may evaluate whether a car assumption is reasonable for a particular data collecting or observational process. This question is complicated by the fact that several distinct versions of mar/car assumptions exist. We therefore first provide an overview over these different versions, in which we highlight the distinction between distributional and coarsening variable induced versions. We show that distributional versions are less restrictive and sufficient for most applications. We then address from two different perspectives the question of when the mar/car assumption is warranted. First we provide a "static" analysis that characterizes the admissibility of the car assumption in terms of the support structure of the joint probability distribution of complete data and incomplete observations. Here we obtain an equivalence characterization that improves and extends a recent result by Grünwald and Halpern. We then turn to a "procedural" analysis that characterizes the admissibility of the car assumption in terms of procedural models for the actual data (or observation) generating process. The main result of this analysis is that the stronger coarsened completely at random (ccar) condition is arguably the most reasonable assumption, as it alone corresponds to data coarsening procedures that satisfy a natural robustness property.

## 1. Introduction

Probabilistic models have become the preeminent tool for reasoning under uncertainty in AI. A probabilistic model consists of a state space $W$, and a probability distribution over the states $x \in W$. A given probabilistic model is used for probabilistic inference based on observations. An observation determines a subset $U$ of $W$ that the true state now is known to belong to. Probabilities then are updated by conditioning on $U$.

The required probabilistic models are often learned from empirical data using statistical parameter estimation techniques. The data can consist of sampled exact states from $W$, but more often it consists of incomplete observations, which only establish that the exact data point $x$ belongs to a subset $U \subseteq W$. Both when learning a probabilistic model, and when using it for probabilistic inference, one should, in principle, distinguish the event that a certain observation $U$ has been made ("$U$ is observed") from the event that the true state of $W$ is a member of $U$ ("$U$ has occurred"). Ignoring this distinction in probabilistic inference can lead to flawed probability assignments by conditioning. Illustrations for this are given by well-known probability puzzles like the Monty-Hall problem or the three prisoners paradox. Ignoring this distinction in statistical learning can lead to the construction of models that do not fit the true distribution on $W$. In spite of these known difficulties, one usually tries to avoid the extra complexity incurred by making the proper distinction between "$U$ is observed" and "$U$ has occurred". In statistics there exists a sizable literature on "ignorability" conditions that permit learning procedures to ignore this distinction. In the AI literature dealing with probabilistic inference this topic has received rather scant attention, though it has been realized early on (Shafer, 1985; Pearl, 1988). Recently, however, Grünwald and Halpern (2003) have provided a more in-depth analysis of ignorability from a probabilistic inference point of view.

The ignorability conditions required for learning and inference have basically the same mathematical form, which is expressed in the *missing at random (mar)* or *coarsened at random (car)* conditions. In this paper we investigate several questions relating to these formal conditions. The central theme of this investigation is to provide a deeper insight into what makes an observational process satisfy, or violate, the coarsened at random condition. This question is studied from two different angles: first (Section 3) we identify qualitative properties of the joint distribution of true states and observations that make the *car* assumption feasible at all. The qualitative properties we here consider are constraints on what states and observations have nonzero probabilities. This directly extends the work of Grünwald and Halpern (2003) (henceforth also referred to as *GH*). In fact, our main result in Section 3 is an extension and improvement over one of the main results in *GH*. Secondly (Section 4), we investigate general types of observational procedures that will lead to *car* observations. This, again, directly extends some of the material in *GH*, as well as earlier work by Gill, van der Laan & Robins (1997) (henceforth also referred to as *GvLR*). We develop a formal framework that allows us to analyze previous and new types of procedural models in a unified and systematic way. In particular, this framework allows us to specify precise conditions for what makes certain types of observational processes "natural" or "reasonable". The somewhat surprising result of this analysis is that the arguably most natural classes of observational processes correspond exactly to those processes that will result in observations that are *coarsened completely at random (ccar)* – a strengthened version of *car* that often has been considered an unrealistically strong assumption.

## 2. Fundamentals of Coarse Data and Ignorability

There exist numerous definitions in the literature of what it means that data is missing or coarsened at random (Rubin, 1976; Dawid & Dickey, 1977; Heitjan & Rubin, 1991; Heitjan, 1994, 1997; Gill et al., 1997; Grünwald & Halpern, 2003). While all capture the same basic principle, various definitions are subtly different in a way that can substantially affect their implications. In Section 2.1 we give a fairly comprehensive overview of the variant definitions, and analyze their relationships. In this survey we aim at providing a uniform framework and terminology for different *mar/car* variants. Definitions are attributed to those earlier sources where their basic content has first appeared, even though our definitions and our terminology can differ in some details from the original versions (cf. also the remarks at the end of Section 2.1).

Special emphasis is placed on the distinction between *distributional* and *coarsening variable induced* versions of *car*. In this paper the main focus will then be on distributional versions. In Section 2.2 we summarize results showing that distributional *car* is sufficient to establish ignorability for probabilistic inference.

## 2.1 Defining *Car*

We begin with the concepts introduced by Rubin (1976) for the special case of data with missing values. Assume that we are concerned with a multivariate random variable $\boldsymbol{X} = (X_1, \ldots, X_k)$, where each $X_i$ takes values in a finite state space $V_i$. Observations of $\boldsymbol{X}$ are incomplete, i.e. we observe values $\boldsymbol{y} = (y_1, \ldots, y_k)$, where each $y_i$ can be either the value $x_i \in V_i$ of $X_i$, or the special 'missingness symbol' $*$. One can view $\boldsymbol{y}$ as the realization of a random variable $\boldsymbol{Y}$ that is a function of $\boldsymbol{X}$ and a *missingness indicator*, $\boldsymbol{M}$, which is a random variable with values in $\{0, 1\}^k$:

$$\boldsymbol{Y} = f(\boldsymbol{X}, \boldsymbol{M}), \tag{1}$$

where $\boldsymbol{y} = f(\boldsymbol{x}, \boldsymbol{m})$ is defined by

$$y_i = \begin{cases} x_i & \text{if } m_i = 0 \\ * & \text{if } m_i = 1 \end{cases}. \tag{2}$$

Rubin's (1976) original definition of missing at random is a condition on the conditional distribution of $\boldsymbol{M}$: the data is *missing at random* iff for all $\boldsymbol{y}$ and all $\boldsymbol{m}$:

$$P(\boldsymbol{M} = \boldsymbol{m} \mid \boldsymbol{X}) \quad \text{is constant on} \quad \{\boldsymbol{x} \mid P(\boldsymbol{X} = \boldsymbol{x}) > 0, f(\boldsymbol{x}, \boldsymbol{m}) = \boldsymbol{y}\}. \tag{3}$$

We refer to this condition as the $\boldsymbol{M}$-*mar* condition, to indicate the fact that it is expressed in terms of the missingness indicator $\boldsymbol{M}$.

**Example 2.1** *Let* $\boldsymbol{X} = (X_1, X_2)$ *with* $V_1 = V_2 = \{p, n\}$. *We interpret* $X_1, X_2$ *as two medical tests with possible outcomes* positive *or* negative. *Suppose that test* $X_1$ *always is performed first on a patient, and that test* $X_2$ *is performed if and only if* $X_1$ *comes out positive. Possible observations that can be made then are*

$$\begin{aligned} (n, *) &= f((n, n), (0, 1)) = f((n, p), (0, 1)), \\ (p, n) &= f((p, n), (0, 0)), \\ (p, p) &= f((p, p), (0, 0)). \end{aligned}$$

*For* $\boldsymbol{y} = (n, *)$ *and* $\boldsymbol{m} = (0, 1)$ *we obtain*

$$P(\boldsymbol{M} = \boldsymbol{m} \mid \boldsymbol{X} = (n, n)) = P(\boldsymbol{M} = \boldsymbol{m} \mid \boldsymbol{X} = (n, p)) = 1,$$

*so that (3) is satisfied. For other values of* $\boldsymbol{y}$ *and* $\boldsymbol{m}$ *condition (3) trivially holds, because the sets of* $\boldsymbol{x}$-*values in (3) then are singletons (or empty).*

We can also eliminate the random vector $\boldsymbol{M}$ from the definition of *mar*, and formulate a definition directly terms of the joint distribution of $\boldsymbol{Y}$ and $\boldsymbol{X}$. For this, observe that each observed $\boldsymbol{y}$ can be identified with the set

$$U(\boldsymbol{y}) := \{\boldsymbol{x} \mid \text{for all } i : y_i \neq * \Rightarrow x_i = y_i\}. \tag{4}$$

The set $U(\boldsymbol{y})$ contains the complete data values consistent with the observed $\boldsymbol{y}$. We can now rephrase $\boldsymbol{M}$-*mar* as

$$P(\boldsymbol{Y} = \boldsymbol{y} \mid \boldsymbol{X}) \quad \text{is constant on} \quad \{\boldsymbol{x} \mid P(\boldsymbol{X} = \boldsymbol{x}) > 0, \boldsymbol{x} \in U(\boldsymbol{y})\}. \tag{5}$$

We call this the *distributional mar* condition, abbreviated *d-mar*, because it is in terms of the joint distribution of the complete data $\boldsymbol{X}$, and the observed data $\boldsymbol{Y}$.

**Example 2.2** *(continued from Example 2.1) We have*

$$U((n,*)) = \{(n,n),(n,p)\}, \ U((p,n)) = \{(p,n)\}, \ U((p,p)) = \{(p,p)\}.$$

*Now we compute*

$$P(\boldsymbol{Y} = (n,*) \mid \boldsymbol{X} = (n,n))) = P(\boldsymbol{Y} = (n,*) \mid \boldsymbol{X} = (n,p))) = 1.$$

*Together with the (again trivial) conditions for the two other possible $\boldsymbol{Y}$-values, this shows (5).*

$\boldsymbol{M}$-*mar* and *d-mar* are equivalent, because given $\boldsymbol{X}$ there is a one-to-one correspondence between $\boldsymbol{M}$ and $\boldsymbol{Y}$, i.e. there exists a function $h$ such that for all $\boldsymbol{x}, \boldsymbol{y}$ with $\boldsymbol{x} \in U(\boldsymbol{y})$:

$$\boldsymbol{y} = f(\boldsymbol{x}, \boldsymbol{m}) \ \Leftrightarrow \ \boldsymbol{m} = h(\boldsymbol{y}) \tag{6}$$

($h$ simply translates $\boldsymbol{y}$ into a $\{0, 1\}$-vector by replacing occurrences of $*$ with 1, and all other values in $\boldsymbol{y}$ with 0). Using (6) one can easily derive a one-to-one correspondence between conditions (3) and (5), and hence obtain the equivalence of $\boldsymbol{M}$-*mar* and *d-mar*.

One advantage of $\boldsymbol{M}$-*mar* is that it easily leads to the strengthened condition of *missing completely at random* (Rubin, 1976):

$$P(\boldsymbol{M} = \boldsymbol{m} \mid \boldsymbol{X}) \quad \text{is constant on} \quad \{\boldsymbol{x} \mid P(\boldsymbol{X} = \boldsymbol{x}) > 0\}. \tag{7}$$

We refer to this as the $\boldsymbol{M}$-*mcar* condition.

**Example 2.3** *(continued from Example 2.2) We obtain*

$$P(\boldsymbol{M} = (0,1) \mid \boldsymbol{X} = (n,p)) = 1 \neq 0 = P(\boldsymbol{M} = (0,1) \mid \boldsymbol{X} = (p,p)).$$

*Thus, the observations here are not $\boldsymbol{M}$-mcar.*

A distributional version of *mcar* is slightly more complex, and we defer its statement to the more general case of coarse data, which we now turn to.

Missing attribute values are only one special way in which observations can be incomplete. Other possibilities include imperfectly observed values (e.g. $X_i$ is only known to be either $x \in V_i$ or $x' \in V_i$), partly attributed values (e.g. for $x \in V_i = V_j$ it is only known that $X_i = x$ or $X_j = x$), etc. In all cases, the incomplete observation of $\boldsymbol{X}$ defines the set of possible instantiations of $\boldsymbol{X}$ that are consistent with the observation. This leads to the general concept of *coarse data* (Heitjan & Rubin, 1991), which generalizes the concept of missing data to observations of arbitrary subsets of the state space. In this general setting it is convenient to abstract from the particular structure of the state space as a product $\times_{i=1}^{k} V_i$ induced by a multivariate $\boldsymbol{X}$, and instead just assume a univariate random variable $X$ taking values in a set $W = \{x_1, \ldots, x_n\}$ (of course, this does not preclude the possibility that in fact $W = \times_{i=1}^{k} V_i$). Abstracting from the missingness indicator $\boldsymbol{M}$, one can imagine coarse data as being produced by $X$ and a *coarsening variable* $G$. Again, one can also take the coarsening variable $G$ out of the picture, and model coarse data directly as the joint distribution of $X$ and a random variable $Y$ (the *observed data*) with values in $2^W$. This is the view we will mostly adopt, and therefore the motivation for the following definition.

**Definition 2.4** *Let $W = \{x_1, \ldots, x_n\}$. The* coarse data space *for $W$ is*

$$\Omega(W) := \{(x, U) \mid x \in W, U \subseteq W : x \in U\}.$$

*A* coarse data distribution *is any probability distribution $P$ on $\Omega(W)$.*

A coarse data distribution can be seen as the joint distribution $P(X, Y)$ of a random variable $X$ with values in $W$, and a random variable $Y$ with values in $2^W \setminus \emptyset$. The joint distribution of $X$ and $Y$ is constrained by the condition $X \in Y$. Note that, thus, coarse data spaces and coarse data distributions actually represent both the true complete data and its coarsened observation. In the remainder of this paper, $P$ without any arguments will always denote a coarse data distribution in the sense of Definition 2.4, and can be used interchangeably with $P(X, Y)$. When we need to refer to (joint) distributions of other random variables, then these are listed explicitly as arguments of $P$. E.g.: $P(X, G)$ is the joint distribution of $X$ and $G$.

*Coarsening variables* as introduced by the following definition are a means for specifying the conditional distribution of $Y$ given $X$.

**Definition 2.5** *Let $G$ be a random variable with values in a finite state space $\Gamma$, and*

$$f : W \times \Gamma \to 2^W \setminus \emptyset, \tag{8}$$

*such that*

- *for all $x$ with $P(X = x) > 0$: $x \in f(x, g)$;*

- *for all $x, x'$ with $P(X = x) > 0, P(X = x') > 0$, all $U \in 2^W \setminus \emptyset$, and all $g \in \Gamma$:*

$$f(x, g) = U, x' \in U \ \Rightarrow \ f(x', g) = U. \tag{9}$$

*We call the pair $(G, f)$ a* coarsening variable *for $X$. Often we also refer to $G$ alone as a coarsening variable, in which case the function $f$ is assumed to be implicitly given.*

*A coarse data distribution $P$ is* induced *by $X$ and $(G, f)$ if $P$ is the joint distribution of $X$ and $f(X, G)$.*

The condition (9) has not always been made explicit in the introduction of coarsening variables. However, as noted by Heitjan (1997), it is usually implied in the concept of a coarsening variable. *GvLR* (pp. 283-285) consider a somewhat more general setup in which $f(x, g)$ does not take values in $2^W$ directly, but $y = f(x, g)$ is some observable from which $U = \alpha(y) \subseteq W$ is obtained via a further mapping $\alpha$. The introduction of such an intermediate observable $Y$ is necessary, for example, when dealing with real-valued random variables $X$. Since we then will not have any statistically tractable models for general distributions on $2^{\mathbb{R}}$, a parameterization $Y$ for a small subset of $2^{\mathbb{R}}$ is needed. For example, $Y$ could take values in $\mathbb{R} \times \mathbb{R}$, and $\alpha(y_1, y_2)$ might be defined as the interval $[min\{y_1, y_2\}, \max\{y_1, y_2\}]$. *GvLR* do not require (9) in general; instead they call $f$ *Cartesian* when (9) is satisfied.

The following definition generalizes property (6) of missingness indicators to arbitrary coarsening variables.

**Definition 2.6** *The coarsening variable* $(G, f)$ *is called* invertible *if there exists a function*

$$h : 2^W \setminus \emptyset \to \Gamma, \tag{10}$$

*such that for all $x, U$ with $x \in U$, and all $g \in \Gamma$:*

$$U = f(x, g) \iff g = h(U). \tag{11}$$

An alternative reading of (11) is that $G$ is observable: from the coarse observation $U$ the value $g \in \Gamma$ can be reconstructed, so that $G$ can be treated as a fully observable random variable.

We can now generalize the definition of missing (completely) at random to the coarse data setting. We begin with the generalization of $\boldsymbol{M}$-*mar*.

**Definition 2.7** (Heitjan, 1997) *Let $G$ be a coarsening variable for $X$. The joint distribution* $P(X, G)$ *is* $G$-car *if for all $U \subseteq W$, and $g \in \Gamma$:*

$$P(G = g \mid X) \quad \text{is constant on} \quad \{x \mid P(X = x) > 0, f(x, g) = U\}. \tag{12}$$

By marginalizing out the coarsening variable $G$ (or by not assuming a variable $G$ in the first place), we obtain the following distributional version of *car*.

**Definition 2.8** (Heitjan & Rubin, 1991) *Let $P$ be a coarse data distribution. $P$ is $d$-car if for all $U \subseteq W$*

$$P(Y = U \mid X) \quad \text{is constant on} \quad \{x \mid P(X = x) > 0, x \in U\}. \tag{13}$$

If $X$ is multivariate, and incompleteness of observations consists of missing values, then $d$-*car* coincides with $d$-*mar*, and $\boldsymbol{M}$-*car* with $\boldsymbol{M}$-*mar*.

Condition (12) refers to the joint distribution of $X$ and $G$, condition (13) to the joint distribution of $X$ and $Y$. Since $Y$ is a function of $X$ and $G$, one can always determine from the joint distribution of $X$ and $G$ whether $d$-*car* holds for their induced coarse data distribution. Conversely, when only the coarse data distribution $P(X, Y)$ and a coarsening variable $G$ inducing $P(X, Y)$ are given, it is in general not possible to determine whether $P(X, G)$ is $G$-*car*, because the joint distribution $P(X, G)$ cannot be reconstructed from the given information. However, under suitable assumptions on $G$ it is possible to infer that $P(X, G)$ is $G$-*car* only from the induced $P(X, Y)$ being $d$-*car*. With the following two theorems we clarify these relationships between $G$-*car* and $d$-*car*. These theorems are essentially restatements in our conceptual framework of results already given by *GvLR* (pp. 284-285).

**Theorem 2.9** *A coarse data distribution $P(X, Y)$ is d-car iff there exists a coarsening variable $G$ inducing $P(X, Y)$, such that $P(X, G)$ is $G$-car.*

**Proof:** First assume that $P(X, Y)$ is *d-car*. We construct a canonical coarsening variable $G$ inducing $P(X, Y)$ as follows: let $\Gamma = 2^W \setminus \emptyset$ and $f(x, U) := U$ for all $x \in W$ and $U \in \Gamma$. Define a $\Gamma$-valued coarsening variable $G$ by $P(G = U \mid X = x) := P(Y = U \mid X = x)$. Clearly, the coarse data distribution induced by $G$ is the original $P(X, Y)$, and $P(X, G)$ is *G-car*.

Conversely, assume that $P(X, G)$ is *G-car* for some $G$ inducing $P(X, Y)$. Let $U \subseteq W$, $x \in U$. Then

$$P(Y = U \mid X = x) = P(G \in \{g \in \Gamma \mid f(x, g) = U\} \mid X = x)$$
$$= \sum_{g \in \Gamma : f(x,g) = U} P(G = g \mid X = x).$$

Because of (9) the summation here is over the same values $g \in \Gamma$ for all $x \in U$. Because of *G-car*, the conditional probabilities $P(G = g \mid X = x)$ are constant for $x \in U$. Thus $P(Y = U \mid X = x)$ is constant for $x \in U$, i.e. *d-car* holds. $\qquad\square$

The following example shows that *d-car* does not in general imply *G-car*, and that a fixed coarse data distribution $P(X, Y)$ can be induced both by a coarsening variable for which *G-car* holds, and by another coarsening variable for which *G-car* does not hold.

**Example 2.10** *(continued from Example 2.3) We have already seen that the coarse data distribution here is d-mar and $\boldsymbol{M}$-mar, and hence d-car and $\boldsymbol{M}$-car.*

*$\boldsymbol{M}$ is not the only coarsening variable inducing $P(X, Y)$. In fact, it is not even the simplest: let $G_1$ be a trivial random variable that can only assume one state, i.e. $\Gamma_1 = \{g\}$. Define $f_1$ by*

$$f_1((n, n), g) = f_1((n, p), g) = \{(n, n), (n, p)\},$$
$$f_1((p, n), g) = \{(p, n)\},$$
$$f_1((p, p), g) = \{(p, p)\}.$$

*Then $G_1$ induces $P(X, Y)$, and $P(X, G_1)$ also is trivially G-car.*

*Finally, let $G_2$ be defined by $\Gamma_2 = \{g_1, g_2\}$ and $f_2(x, g_i) = f_1(x, g)$ for all $x \in W$ and $i = 1, 2$. Thus, $G_2$ is just like $G_1$, but the trivial state space of $G_1$ has been split into two elements with identical meaning. Let the conditional distribution of $G_2$ given $X$ be*

$$P(G_2 = g_1 \mid X = (n, n)) = P(G_2 = g_2 \mid X = (n, p)) = 2/3,$$
$$P(G_2 = g_2 \mid X = (n, n)) = P(G_2 = g_1 \mid X = (n, p)) = 1/3,$$
$$P(G_2 = g_1 \mid X = (p, n)) = P(G_2 = g_1 \mid X = (p, p)) = 1.$$

*Again, $G_2$ induces $P(X, Y)$. However, $P(X, G_2)$ is not G-car, because*

$$f_2((n, n), g_1) = f_2((n, p), g_1) = \{(n, n), (n, p)\},$$
$$P(G_2 = g_1 \mid X = (n, n)) \neq P(G_2 = g_1 \mid X = (n, p))$$

*violates the G-car condition. $G_2$ is not invertible in the sense of Definition 2.6: when, for example, $U = \{(n, n), (n, p)\}$ is observed, it is not possible to determine whether the value of $G_2$ was $g_1$ or $g_2$.*

The following theorem shows that the non-invertibility of $G_2$ in the preceding example is the reason why we cannot deduce *G-car* for $P(X, G_2)$ from the *d-car* property of the induced $P(X, Y)$. This theorem completes our picture of the *G-car*/ *d-car* relationship.

**Theorem 2.11** *Let $P(X, Y)$ be a coarse data distribution, $G$ an invertible coarsening variable inducing $P(X, Y)$. If $P(X, Y)$ is d-*car*, then $P(X, G)$ is G-*car*.*

**Proof:** Let $U \subseteq W$, $g \in \Gamma$, and $x \in U$, such that $P(X = x) > 0$ and $f(x, g) = U$. Since $G$ is invertible, we have that $f(x, g') \neq U$ for all $g' \neq g$, and hence

$$P(G = g \mid X = x) = P(Y = U \mid X = x).$$

From the assumption that $P$ is *d-car* it follows that the right-hand probability is constant for $x \in U$, and hence the same holds for the left-hand side, i.e. *G-car* holds.  □

We now turn to coarsening completely at random (*ccar*). It is straightforward to generalize the definition of $\boldsymbol{M}$-*mcar* to general coarsening variables:

**Definition 2.12** (Heitjan, 1994) *Let $G$ be a coarsening variable for $X$. The joint distribution $P(X, G)$ is G-*ccar* if for all $g \in \Gamma$*

$$P(G = g \mid X) \quad \text{is constant on} \ \ \{x \mid P(X = x) > 0\}. \tag{14}$$

A distributional version of *ccar* does not seem to have been formalized previously in the literature. *GvLR* refer to coarsening completely at random, but do not provide a formal definition. However, it is implicit in their discussion that they have in mind a slightly restricted version of our following definition (the restriction being a limitation to the case $k = 1$ in Theorem 2.14 below).

We first observe that one cannot give a definition of *d-ccar* as a variant of Definition 2.12 in the same way as Definition 2.8 varies Definition 2.7, because that would lead us to the condition that $P(Y = U \mid X)$ is constant on $\{x \mid P(X = x) > 0\}$. This would be inconsistent with the existence of $x \in W \setminus U$ with $P(X = x) > 0$. However, the real semantic core of *d-car*, arguably, is not so much captured by Definition 2.8, as by the characterization given in Theorem 2.9. For *d-ccar*, therefore, we make an analogous characterization the basis of the definition:

**Definition 2.13** *A coarse data distribution $P(X, Y)$ is d-*ccar* iff there exists a coarsening variable $G$ inducing $P(X, Y)$, such that $P(X, G)$ is G-*ccar*.*

The following theorem provides a constructive characterization of *d-ccar*.

**Theorem 2.14** *A coarse data distribution $P(X, Y)$ is d-*ccar* iff there exists a family $\{\mathcal{W}_1, \dots, \mathcal{W}_k\}$ of partitions of $W$, and a probability distribution $(\lambda_1, \dots, \lambda_k)$ on $(\mathcal{W}_1, \dots, \mathcal{W}_k)$, such that for all $x \in W$ with $P(X = x) > 0$:*

$$P(Y = U \mid X = x) = \sum_{\substack{i \in 1, \dots, k \\ x \in U \in \mathcal{W}_i}} \lambda_i. \tag{15}$$

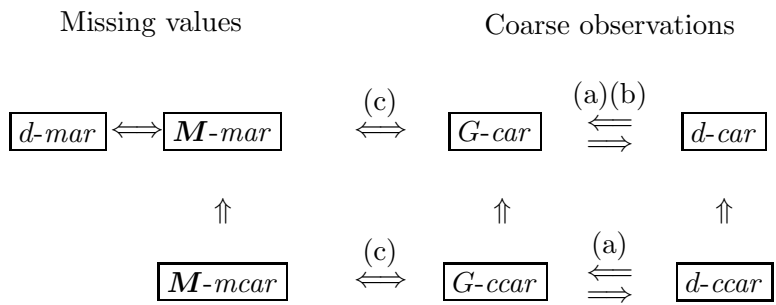

Figure 1: Versions of *car*. (a): there exists $G$ such that this implication holds; (b): for all invertible $G$ this implication holds; (c): equivalence holds for $G = \boldsymbol{M}$.

**Proof:** Assume that $P$ is *d-ccar*. Let $G$ be a coarsening variable inducing $P(X,Y)$, such that $P(X,G)$ is *G-ccar*. Because of (9), each value $g_i \in \Gamma$ induces a partition $\mathcal{W}_i = \{U_{i,1}, \ldots, U_{i,k(i)}\}$, such that $f(x, g_i) = U_{i,j} \Leftrightarrow x \in U_{i,j}$. The partitions $\mathcal{W}_i$ together with $\lambda_i := P(G = g_i \mid X)$ then provide a representation of $P(X,Y)$ in the form (15).

Conversely, if $P(X,Y)$ is given by (15) via partitions $\mathcal{W}_1, \ldots, \mathcal{W}_k$ and parameters $\lambda_i$, one defines a coarsening variable $G$ with $\Gamma = \{1, \ldots, k\}$, $P(G = g_i \mid X = x) = \lambda_i$ for all $x$ with $P(X = x) > 0$, and $f(x, i)$ as that $U \in \mathcal{W}_i$ containing $x$. $P(X, G)$ then is *G-ccar* and induces $P(X,Y)$, and hence $P(X,Y)$ is *d-ccar*. $\square$

As before, we have that the *G-ccar* property of $P(X,G)$ cannot be determined from the induced coarse data distribution:

**Example 2.15** *(continuation of Example 2.10)* $P(X,Y)$ *is d-*ccar *and induced by any of the three coarsening variables* $\boldsymbol{M}$, $G_1$, $G_2$. *However,* $P(X,G_1)$ *is G-*ccar*, while* $P(X, \boldsymbol{M})$ *and* $P(X,G_2)$ *are not.*

The previous example also shows that no analog of Theorem 2.11 holds for *ccar*: $\boldsymbol{M}$ is invertible, but from *d-ccar* for the induced $P(X,Y)$ we here cannot infer *G-ccar* for $P(X, \boldsymbol{M})$.

Figure 1 summarizes the different versions of *mar/car* we have considered. The distributional versions *d-car* and *d-ccar* are weaker than their $\boldsymbol{M}$- and $G$- counterparts, and therefore the less restrictive assumptions. At the same time, they are sufficient to establish ignorability for most statistical learning and probabilistic inference tasks. For the case of probabilistic inference this will be detailed by Theorem 2.18 in the following section. For statistical inference problems, too, the required ignorability results can be obtained from the distributional *car* versions, unless a specific coarsening variable is explicitly part of the inference problem. Whenever a coarsening variable $G$ is introduced only as an artificial construct for modeling the connection between incomplete observations and complete data, one must be aware that the *G-car* and *G-ccar* conditions can be unnecessarily restrictive, and may lead us to reject ignorability when, in fact, ignorability holds (cf. Examples 2.3 and 2.15).

We conclude this section with three additional important remarks on definitions of *car*, which are needed to complete the picture of different approaches to *car* in the literature:

*Remark 1:* All of the definitions given here are *weak* versions of *mar/car*. Corresponding *strong* versions are obtained by dropping the restriction $P(X = x) > 0$ from (3),(5),(7),(12),(13),(14), respectively (15). Differences between weak and strong versions of *car* are studied in previous work (Jaeger, 2005). The results there obtained indicate that in the context of probability updating the weak versions are more suitable. For this reason we do not go into the details of strong versions here.

*Remark 2:* Our definitions of *car* differ from those originally given by Rubin and Heitjan in that our definitions are "global" definitions that view *mar/car* as a property of a joint distribution of complete and coarse data. The original definitions, on the other hand, are conditional on a single observation $Y = U$, and do not impose constraints on the joint distribution of $X$ and $Y$ for other values of $Y$. These "local" *mar/car* assumptions are all that is required to justify the application of certain probabilistic or statistical inference techniques to the single observation $Y = U$. The global *mar/car* conditions we stated justify these inference techniques as general strategies that would be applied to any possible observation. Local versions of *car* are more natural under a Bayesian statistical philosophy, whereas global versions are required under a frequentist interpretation. Global versions of *car* have also been used in other works (e.g., Jacobsen & Keiding, 1995; Gill et al., 1997; Nielsen, 1997; Cator, 2004).

*Remark 3:* The definitions and results stated here are strictly limited to the case of finite $W$. As already indicated in the discussion following Definition 2.5, extensions of *car* to more general state spaces $C$ typically require a setup in which observations are modeled by a random variable taking values in a more manageable state space than $2^C$. Several such formalizations of *car* for continuous state spaces have been investigated (e.g., Jacobsen & Keiding, 1995; Gill et al., 1997; Nielsen, 2000; Cator, 2004).

## 2.2 Ignorability

*Car* and *mar* assumptions are needed for ignoring the distinction between "$U$ is observed" and "$U$ has occurred" in statistical inference and probability updating. In statistical inference, for example, *d-car* is required to justify likelihood maximizing techniques like the *EM* algorithm (Dempster, Laird, & Rubin, 1977) for learning from incomplete data. In this paper the emphasis is on probability updating. We therefore briefly review the significance of *car* in this context. We use the well-known Monty Hall problem.

**Example 2.16** *A contestant at a game show is asked to choose one from three closed doors $A, B, C$, behind one of which is hidden a valuable prize, the others each hiding a goat. The contestant chooses door $A$, say. The host now opens door $B$, revealing a goat. At this point the contestant is allowed to change her choice from $A$ to $C$. Would this be advantageous?*

*Being a savvy probabilistic reasoner, the contestant knows that she should analyze the situation using the coarse data space $\Omega(\{A, B, C\})$, and compute the probabilities*

$$P(X = A \mid Y = \{A, C\}), \ P(X = C \mid Y = \{A, C\}).$$

*She makes the following assumptions: 1. A-priori all doors are equally likely to hide the prize. 2. Independent from the contestants choice, the host will always open one door. 3. The host will never open the door chosen by the contestant. 4. The host will never open the door hiding the prize. 5. If more than one possible door remain for the host, he will determine by a fair coin flip which one to open. From this, the contestant first obtains*

$$P(Y = \{A, C\} \mid X = A) = 1/2, \ \ P(Y = \{A, C\} \mid X = C) = 1, \tag{16}$$

*and then*

$$P(X = A \mid Y = \{A, C\}) = 1/3, \ \ P(X = C \mid Y = \{A, C\}) = 2/3.$$

*The conclusion, thus, is that it will be advantageous to switch to door C. A different conclusion is obtained by simply conditioning in the state space W on "$\{A, C\}$ has occurred":*

$$P(X = A \mid X \in \{A, C\}) = 1/2, \ \ P(X = C \mid X \in \{A, C\}) = 1/2.$$

**Example 2.17** *Consider a similar situation as in the previous example, but now assume that just after the contestant has decided for herself that she would pick door A, but before communicating her choice to the host, the host says "let me make things a little easier for you", opens door B, and reveals a goat. Would changing from A to C now be advantageous?*

*The contestant performs a similar analysis as before, but now based on the following assumptions: 1. A-priori all doors are equally likely to hide the prize. 2. The host's decision to open a door was independent from the location of the prize. 3. Given his decision to open a door, the host chose by a fair coin flip one of the two doors not hiding the prize. Now*

$$P(Y = \{A, C\} \mid X = A) = P(Y = \{A, C\} \mid X = C), \tag{17}$$

*and hence*

$$P(X = A \mid Y = \{A, C\}) = 1/2, \ \ P(X = C \mid Y = \{A, C\}) = 1/2.$$

*In particular here*

$$P(X = A \mid Y = \{A, C\}) = P(X = A \mid X \in \{A, C\})$$
$$P(X = C \mid Y = \{A, C\}) = P(X = C \mid X \in \{A, C\}),$$

*i.e. the difference between "$\{A, C\}$ is observed" and "$\{A, C\}$ has occurred" can be ignored for probability updating.*

The coarse data distribution in Example 2.16 is not *d-car* (as evidenced by (16)), whereas the coarse data distribution in Example 2.17 is *d-car* (as shown, in part, by (17)). The connection between ignorability in probability updating and the *d-car* assumption has been shown in *GvLR* and *GH*. The following theorem restates this connection in our terminology.

**Theorem 2.18** *Let P be a coarse data distribution. The following are equivalent:*

**(i)** *P is d-*car.

**(ii)** *For all $x \in W$, $U \subseteq W$ with $x \in U$ and $P(Y = U) > 0$:*

$$P(X = x \mid Y = U) = P(X = x \mid X \in U).$$

**(iii)** *For all $x \in W$, $U \subseteq W$ with $x \in U$ and $P(X = x) > 0$:*

$$P(Y = U \mid X = x) = \frac{P(Y = U)}{P(X \in U)}.$$

The equivalence (i)⇔(ii) is shown in *GH*, based on *GvLR*. For the equivalence with (iii) see (Jaeger, 2005).

## 3. Criteria for *Car* and *Ccar*

Given a coarse data distribution $P$ it is, in principle, easy to determine whether $P$ is *d-car* (*d-ccar*) based on Definition 2.8, respectively Theorem 2.14 (though in case of *d-ccar* a test might require a search over possible families of partitions). However, typically $P$ is not completely known. Instead, we usually have some partial information about $P$. In the case of statistical inference problems this information consists of a sample $U_1, \ldots, U_N$ of the coarse data variable $Y$. In the case of conditional probabilistic inference, we know the marginal of $P$ on $W$. In both cases we would like to decide whether the partial knowledge of $P$ that we possess, in conjunction with certain other assumptions on the structure of $P$ that we want to make, is consistent with *d-car*, respectively *d-ccar*, i.e. whether there exists a distribution $P$ that is *d-car* (*d-ccar*), and satisfies our partial knowledge and our additional assumptions.

In statistical problems, additional assumptions on $P$ usually come in the form of a parametric representation of the distribution of $X$. When $\boldsymbol{X} = (X_1, \ldots, X_k)$ is multivariate, such a parametric representation can consist, for example, in a factorization of the joint distribution of the $X_i$, as induced by certain conditional independence assumptions. In probabilistic inference problems an analysis of the evidence gathering process can lead to assumptions about the likelihoods of possible observations. In all these cases, one has to determine whether the constraints imposed on $P$ by the partial knowledge and assumptions are consistent with the constraints imposed by the *d-car* assumption. In general, this will lead to computationally very difficult optimization or constraint satisfaction problems.

Like *GH*, we will focus in this section on a rather idealized special problem within this wider area, and consider the case where our constraints on $P$ only establish what values the variables $X$ and $Y$ can assume with nonzero probability, i.e. the constraints on $P$ consist of prescribed sets of support for $X$ and $Y$. We can interpret this special case as a reduced form of a more specific statistical setting, by assuming that the observed sample $U_1, \ldots, U_N$ only is used to infer what observations are possible, and that the parametric model for $X$, too, only is used to determine what $x \in W$ have nonzero probabilities. Similarly, in the probabilistic inference setting, this special case occurs when the knowledge of the distribution of $X$ only is used to identify the $x$ with $P(X = x) > 0$, and assumptions on the evidence generation only pertain to the set of possible observations.

*GH* represent a specific support structure of $P$ in form of a $0,1$-matrix, which they call the "CARacterizing matrix". In the following definition we provide an equivalent, but different, encoding of support structures of $P$.

**Definition 3.1** *A support hypergraph (for a given coarse data space $\Omega(W)$) is a hypergraph of the form $(\mathcal{N}, W')$, where*

- $\mathcal{N} \subseteq 2^W \setminus \emptyset$ *is the set of nodes,*

- $W' \subseteq W$ *is the set of edges, such that each edge* $x \in W'$ *just contains the nodes* $\{U \in \mathcal{N} \mid x \in U\}.$

$(\mathcal{N}, W')$ *is called the* support hypergraph *of the distribution* $P$ *on* $\Omega(W)$ *iff* $\mathcal{N} = \{U \subseteq 2^W \setminus \emptyset \mid P(Y = U) > 0\},$ *and* $W' = \{x \in W \mid P(X = x) > 0\}.$ *A support hypergraph is* car-compatible *iff it is the support hypergraph of some* $d$-car *distribution* $P$.

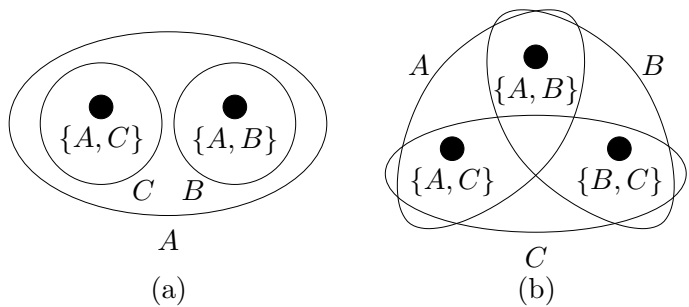

Figure 2: Support hypergraphs for Examples 2.16 and 2.17

**Example 3.2** *Figure 2 (a) shows the support hypergraph of the coarse data distribution in Example 2.16; (b) for Example 2.17.*

The definition of support hypergraph may appear strange, as a much more natural definition would take the states $x \in W$ with $P(X = x) > 0$ as the nodes, and the observations $U \subseteq W$ as the edges. The support hypergraph of Definition 3.1 is just the dual of this natural support hypergraph. It turns out that these duals are more useful for the purpose of our analysis.

A support hypergraph can contain multiple edges containing the same nodes. This corresponds to multiple states that are not distinguished by any of the possible observations. Similarly, a support hypergraph can contain multiple nodes that belong to exactly the same edges. This corresponds to different observations $U, U'$ with $U \cap \{x \mid P(X = x) > 0\} = U' \cap \{x \mid P(X = x) > 0\}$. On the other hand, a support hypergraph cannot contain any node that is not contained in at least one edge (this would correspond to an observation $U$ with $P(Y = U) > 0$ but $P(X \in U) = 0$). Similarly, it cannot contain empty edges. These are the only restrictions on support hypergraphs:

**Theorem 3.3** *A hypergraph* $(\mathcal{N}, \mathcal{E})$ *with finite* $\mathcal{N}$ *and* $\mathcal{E}$ *is the support hypergraph of some distribution* $P$, *iff each node in* $\mathcal{N}$ *is contained in at least one edge from* $\mathcal{E}$, *and all edges are nonempty.*

**Proof:** Let $W = \mathcal{E}$ and define $P(X = x) = 1/|\mathcal{E}|$ for all $x \in W$. For each node $n \in \mathcal{N}$ let $U(n)$ be $\{x \in W \mid n \in x\}$ (nonempty!), and define $P(Y = U(n) \mid X = x) = 1/|x|$. Then $(\mathcal{N}, \mathcal{E})$ is the support hypergraph of $P$. $\qquad \square$

While (almost) every hypergraph, thus, can be the support hypergraph of some distribution, only rather special hypergraphs can be the support hypergraphs of a *d-car* distribution. Our goal, now, is to characterize these *car*-compatible support hypergraphs. The following proposition gives a first such characterization. It is similar to lemma 4.3 in *GH*.

**Proposition 3.4** *The support hypergraph* $(\mathcal{N}, W')$ *is car-compatible iff there exists a function* $\nu : \mathcal{N} \to (0,1]$, *such that for all* $x \in W'$

$$\sum_{U \in \mathcal{N} : U \in x} \nu(U) = 1 \tag{18}$$

**Proof:** First note that in this proposition we are looking at $x$ and $U$ as edges and nodes, respectively, of the support hypergraph, so that writing $U \in x$ makes sense, and means the same as $x \in U$ when $x$ and $U$ are seen as states, respectively sets of states, in the coarse data space.

Suppose $(\mathcal{N}, W')$ is the support hypergraph of a *d-car* distribution $P$. It follows from Lemma 2.18 that $\nu(U) := P(Y = U)/P(X \in U)$ defines a function $\nu$ with the required property. Conversely, assume that $\nu$ is given. Let $P(X)$ be any distribution on $W$ with support $W'$. Setting $P(Y = U \mid X = x) := \nu(U)$ for all $U \in \mathcal{N}$, and $x \in W' \cap U$ extends $P$ to a *d-car* distribution whose support hypergraph is just $(\mathcal{N}, W')$. $\qquad\square$

**Corollary 3.5** *If the support hypergraph contains (properly) nested edges, then it is not car-compatible.*

**Example 3.6** *The support hypergraph from Example 2.16 contains nested edges. Without any numerical computations, it thus follows alone from the qualitative analysis of what observations could have been made, that the coarse data distribution is not d-car, and hence conditioning is not a valid update strategy.*

The proof of Proposition 3.4 shows (as already observed by *GH*) that if a support hypergraph is *car*-compatible, then it is *car*-compatible for any given distribution $P(X)$ with support $W'$, i.e. the support assumptions encoded in the hypergraph, together with the *d-car* assumption (if jointly consistent), do not impose any constraints on the distribution of $X$ (other than having the prescribed set of support). The same is not true for the marginal of $Y$: for a *car*-compatible support hypergraph $(\mathcal{N}, W')$ there will usually also exist distributions $P(Y)$ on $\mathcal{N}$ such that $P(Y)$ cannot be extended to a *d-car* distribution with the support structure specified by the hypergraph $(\mathcal{N}, W')$.

Proposition 3.4 already provides a complete characterization of *car*-compatible support hypergraphs, and can be used as the basis of a decision procedure for *car*-compatibility using methods for linear constraint satisfaction. However, Proposition 3.4 does not provide very much real insight into what makes an evidence hypergraph *car*-compatible. Much more intuitive insight is provided by Corollary 3.5. The criterion provided by Corollary 3.5 is not complete: as the following example shows, there exist support hypergraphs without nested edges that are not *car*-compatible.

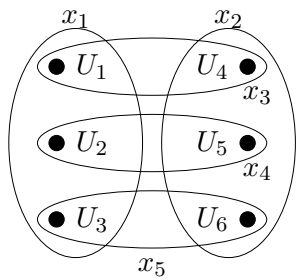

Figure 3: *car*-incompatible support hypergraph without nested edges

**Example 3.7** *Let $(\mathcal{N}, W')$ be as shown in Figure 3. By assuming the existence of a suitable function $\nu$, and summing (18) once over $x_1$ and $x_2$, and once over $x_3, x_4, x_5$, we obtain the contradiction $2 = \sum_{i=1}^{6} \nu(U_i) = 3$. Thus, $(\mathcal{N}, W')$ is not car-compatible.*

We now proceed to extend the partial characterization of *car*-compatibility provided by Corollary 3.5 to a complete characterization. Our following result improves on theorem 4.4 of *GH* by giving a necessary and sufficient condition for *car*-compatibility, rather than just several necessary ones, and, arguably, by providing a criterion that is more intuitive and easier to apply. Our characterization is based on the following definition.

**Definition 3.8** *Let $(\mathcal{N}, W')$ be a support hypergraph. Let $\boldsymbol{x} = x_1, \ldots, x_k$ be a finite sequence of edges from $W'$, possibly containing repetitions of the same edge. Denote the length $k$ of the sequence by $|\boldsymbol{x}|$. For $x \in W$ we denote with $1_x$ the indicator function on $\mathcal{N}$ induced by $x$, i.e.*

$$1_x(U) := \left\{ \begin{array}{ll} 1 & \text{if } U \in x \\ 0 & \text{else} \end{array} \right. \qquad (U \in \mathcal{N}).$$

*The function $1_{\boldsymbol{x}}(U) := \sum_{x \in \boldsymbol{x}} 1_x(U)$ then counts the number of edges in $\boldsymbol{x}$ that contain $U$. For two sequences $\boldsymbol{x}, \boldsymbol{x}'$ we write $1_{\boldsymbol{x}} \leq 1_{\boldsymbol{x}'}$ iff $1_{\boldsymbol{x}}(U) \leq 1_{\boldsymbol{x}'}(U)$ for all $U$.*

**Example 3.9** *For the evidence hypergraph in Figure 3 we have that $1_{(x_1, x_2)} = 1_{(x_3, x_4, x_5)}$ is the function on $\mathcal{N}$ which is constant 1.*

*For $\boldsymbol{x} = (x_1, x_3, x_4, x_5)$ one obtains $1_{\boldsymbol{x}}(U) = 2$ for $U = U_1, U_2, U_3$, and $1_{\boldsymbol{x}}(U) = 1$ for $U = U_4, U_5, U_6$. The same function also is defined by $\boldsymbol{x} = (x_1, x_1, x_2)$.*

*In any evidence hypergraph, one has that for two single edges $x, x'$: $1_x < 1_{x'}$ iff $x$ is a proper subset of $x'$.*

We now obtain the following characterization (which is partly inspired by known conditions for the existence of finitely additive measures, see Bhaskara Rao & Bhaskara Rao, 1983):

**Theorem 3.10** *The support hypergraph $(\mathcal{N}, W')$ is car-compatible iff for every two sequences $\boldsymbol{x}, \boldsymbol{x}'$ of edges from $W'$ we have*

$$1_{\boldsymbol{x}} = 1_{\boldsymbol{x}'} \quad \Rightarrow \quad |\boldsymbol{x}| = |\boldsymbol{x}'|, \text{ and} \tag{19}$$

$$1_{\boldsymbol{x}} \leq 1_{\boldsymbol{x}'}, 1_{\boldsymbol{x}} \neq 1_{\boldsymbol{x}'} \quad \Rightarrow \quad |\boldsymbol{x}| < |\boldsymbol{x}'| . \tag{20}$$

**Proof:** Denote $k := |W'|$, $l := |\mathcal{N}|$. Let $\boldsymbol{A} = (a_{i,j})$ be the incidence matrix of $(\mathcal{N}, W')$, i.e. $\boldsymbol{A}$ is an $k \times l$ matrix with $a_{i,j} = 1$ if $U_j \in x_i$, and $a_{i,j} = 0$ if $U_j \notin x_i$ (using some indexings $i, j$ for $W'$ and $\mathcal{N}$).

Condition (3.4) now reads:

$$\text{exists } \boldsymbol{\nu} \in (0, 1]^l \text{ with } \boldsymbol{A}\boldsymbol{\nu} = \mathbf{1} \tag{21}$$

(here $\mathbf{1}$ is a vector of $k$ ones). An edge indicator function $1_{\boldsymbol{x}}$ can be represented as a row vector $\boldsymbol{z} \in \mathbb{N}^k$, where $z_i$ is the number of times $x_i$ occurs in $\boldsymbol{x}$. Then $1_{\boldsymbol{x}}$ can be written as the row vector $\boldsymbol{z}\boldsymbol{A}$, and the conditions of Theorem 3.10 become: for all $\boldsymbol{z}, \boldsymbol{z}' \in \mathbb{N}^k$:

$$\boldsymbol{z}\boldsymbol{A} = \boldsymbol{z}'\boldsymbol{A} \;\Rightarrow\; \boldsymbol{z} \cdot \mathbf{1} = \boldsymbol{z}' \cdot \mathbf{1}, \tag{22}$$

$$\boldsymbol{z}\boldsymbol{A} \leq \boldsymbol{z}'\boldsymbol{A}, \; \boldsymbol{z}\boldsymbol{A} \neq \boldsymbol{z}'\boldsymbol{A} \Rightarrow\; \boldsymbol{z} \cdot \mathbf{1} < \boldsymbol{z}' \cdot \mathbf{1}. \tag{23}$$

Subtracting right sides, this is equivalent to: for all $\boldsymbol{z} \in \mathbb{Z}^k$:

$$\boldsymbol{z}\boldsymbol{A} = \mathbf{0} \;\Rightarrow\; \boldsymbol{z} \cdot \mathbf{1} = \mathbf{0}, \tag{24}$$

$$\boldsymbol{z}\boldsymbol{A} \leq \mathbf{0}, \; \boldsymbol{z}\boldsymbol{A} \neq \mathbf{0} \Rightarrow\; \boldsymbol{z} \cdot \mathbf{1} < \mathbf{0}. \tag{25}$$

Using Farkas's lemma (see e.g. Schrijver, 1986, Section 7.3), one now obtains that conditions (24) and (25) are necessary and sufficient for (21). For the application of Farkas's lemma to our particular setting one has to observe that since $\boldsymbol{A}$ and $\mathbf{1}$ are rational, it is sufficient to have (24) and (25) for rational $\boldsymbol{z}$ (cf. Schrijver, 1986[p.85]). This, in turn, is equivalent to having (24) and (25) for integer $\boldsymbol{z}$. The strict positivity of the solution $\boldsymbol{\nu}$ can be derived from conditions (24) and (25) by analogous arguments as for Corollary 7.1k in (Schrijver, 1986). $\qquad\square$

**Example 3.11** *From Example 3.9 we immediately obtain that nested edges $x, x'$ violate (20), and hence we again obtain Corollary 3.5. Also the sequences $(x_1, x_2)$ and $(x_3, x_4, x_5)$ of the support hypergraph in Figure 3 violate (19), so that we again obtain the* car*-incompatibility of that hypergraph.*

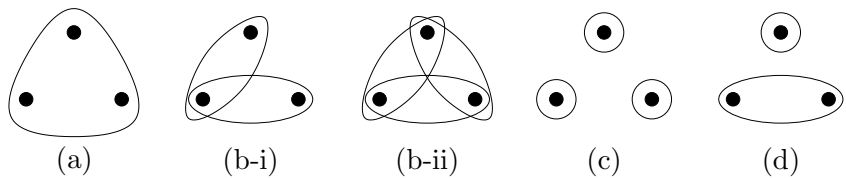

(a)  (b-i)  (b-ii)  (c)  (d)

Figure 4: *Car*-compatible support hypergraphs with three nodes

**Example 3.12** GH *(Example 4.6) derive a complete characterization of* car*-compatibility for the case that exactly three different observations can be made with positive probability. In our framework this amounts to finding all support hypergraphs with three nodes that satisfy the conditions of Theorem 3.10. The possible solutions are shown in Figure 4 (omitting equivalent solutions obtained by duplicating edges). The labeling (a)-(d) of the solutions* *corresponds to the case enumeration in* GH. *It is easy to verify that the shown support hypergraphs all satisfy (19) and (20). That these are the only such hypergraphs follows from the facts that adding a new edge to any of the shown hypergraphs either leads to a hypergraph already on our list (only possible for the pair (b-i) and (b-ii)), or introduces a pair of nested edges. Similarly, deleting any edge either leads to a hypergraph already shown, or to an invalid hypergraph in which not all nodes are covered.*

## 4. Procedural Models

So far we have emphasized the distributional perspective on *car*. We have tried to identify *car* from the joint distribution of complete and coarse data. From this point of view coarsening variables are an artificial construct that is introduced to describe the joint distribution. In some cases, however, a coarsening variable can also model an actual physical, stochastic process that leads to the data coarsening. In such cases, the analysis should obviously take this concrete model for the underlying coarsening process into account. In this section we study *d-car* distributions in terms of such *procedural models* for the data generating mechanism. Our results in this section extend previous investigations of *car mechanisms* in *GvLR* and *GH*.

Our first goal in this section is to determine canonical procedural models for coarsening mechanisms leading to *d-car* data. Such canonical models can be used in practice for evaluating whether a *d-car* assumption is warranted for a particular data set under investigation by matching the (partially known or hypothesized) coarsening mechanism of the data against any of the canonical models. Our investigation will then focus on properties that one may expect reasonable or natural procedural models to possess. These properties will be captured in two formal conditions of *honesty* and *robustness*. The analysis of these conditions will provide new strong support for the *d-ccar* assumption.

The following definition of a procedural model is essentially a generalization of coarsening variables, obtained by omitting condition (9), and by replacing the single variable $G$ with a (potentially infinite) sequence $\boldsymbol{G}$.

**Definition 4.1** *Let $P$ be a coarse data distribution on $\Omega(W)$. A procedural model for $P$ is given by*

- *A random variable $X$ distributed according to the marginal of $P$ on $W$.*

- *A finite or infinite sequence $\boldsymbol{G} = G_1, G_2, \ldots$, of random variables, such that $G_i$ takes values in a finite set $\Gamma_i$ ($i \geq 1$).*

- *A function $f : W \times \times_i \Gamma_i \rightarrow 2^W \setminus \emptyset$, such that $(X, f(X, \boldsymbol{G}))$ is distributed according to $P$.*

We also call a procedural model $(X, \boldsymbol{G}, f)$ a *car* model (*ccar* model), if the coarse data distribution $P$ it defines is *d-car* (*d-ccar*). In the following we denote $\times_i \Gamma_i$ with $\boldsymbol{\Gamma}$.

Some natural coarsening processes are modeled with real-valued coarsening variables (e.g. censoring times, Heitjan & Rubin, 1991). We can accommodate real-valued variables $Z$ in our framework by identifying $Z$ with a sequence of binary random variables $Z_i$ ($i \geq 1$) defining its binary representation. A sequence $\boldsymbol{G}$ containing a continuous $Z = G_i$ can then be replaced with a sequence $\boldsymbol{G'}$ in which the original variables $G_j$ $(j \neq i)$ are interleaved with the binary $Z_i$.

In the following, we will not discuss measurability issues in detail (only in Appendix A will a small amount of measure theory be needed). It should be mentioned, however, that it is always assumed that $W$ and the $\Gamma_i$ are equipped with a $\sigma$-algebra equal to their powerset, that on $W \times \boldsymbol{\Gamma}$ we have the generated product $\sigma$-algebra, and that $f^{-1}(U)$ is measurable for all $U \subseteq W$.

| $f_{2.16}$ | $A$ | $B$ | $C$ | | $f_{2.17}$ | $A$ | $B$ | $C$ |
|---|---|---|---|---|---|---|---|---|
| $h$ | $\{A, C\}$ | $\{A, B\}$ | $\{A, C\}$ | | $h$ | $\{A, C\}$ | $\{B, C\}$ | $\{B, C\}$ |
| $t$ | $\{A, B\}$ | $\{A, B\}$ | $\{A, C\}$ | | $t$ | $\{A, B\}$ | $\{A, B\}$ | $\{A, C\}$ |

Table 1: Procedural models for Examples 2.16 and 2.17

**Example 4.2** *Natural procedural models for the coarse data distributions of Examples 2.16 and 2.17 are constructed by letting $\boldsymbol{G} = F$ represent the coin flip that determines the door to be opened. Suppose that in both examples the host has the following rule for matching the result of the coin flip with potential doors for opening: when the coin comes up heads, then the host opens the door that is first in alphabetical order among all doors (two, at most) that his rules permit him to open. If the coin comes up tails, he opens the door last in alphabetical order. This is formally represented by a procedural model in which $X$ is a $\{A, B, C\}$-valued, uniformly distributed random variable, $\boldsymbol{G}$ consists of a single $\{h, t\}$-valued, uniformly distributed random variable $F$, and $X$ and $F$ are independent.*

*Table 1 completes the specification of the procedural models by defining the functions $f_{2.16}$ and $f_{2.17}$ for the respective examples. Note that neither $(F, f_{2.16})$ nor $(F, f_{2.17})$ are coarsening variables in the sense of Definition 2.5, as e.g. $A \in f_{2.16}(B, h) \neq f_{2.16}(A, h)$, in violation of (9).*

The two procedures described in the preceding example appear quite similar, yet one produces a *d-car* distribution while the other does not. We are now interested in identifying classes of procedural models that are guaranteed to induce *d-car* (*d-ccar*) distributions. Conversely, for any given *d-car* distribution $P$, we would like to identify procedural models that might have induced $P$. We begin with a class of procedural models that stands in a trivial one-to-one correspondence with *d-car* distributions.

**Example 4.3** *(Direct* car *model) Let $X$ be a $W$-valued random variable, and $\boldsymbol{G} = G_1$ with $\Gamma_1 = 2^W \setminus \emptyset$. Let the joint distribution of $X$ and $G_1$ be such that $P(G_1 = U \mid X = x) = 0$ when $x \notin U$, and*

$$P(G_1 = U \mid X = x) \text{ is constant on } \{x \mid P(X = x) > 0, x \in U\}. \tag{26}$$

*Define $f(x, U) = U$. Procedural models of this form are just the coarsening variable representations of d-*car *distributions that we already encountered in Theorem 2.9. Hence, a coarse data distribution $P$ is d-*car *iff it is induced by a direct* car *model.*

The direct *car* models are not much more than a restatement of the *d-car* definition. They do not help us very much in our endeavor to identify canonical observational or data-generating processes that will lead to *d-car* distributions, because the condition (26) does not correspond to an easily interpretable condition on an experimental setup.

For *d-ccar* the situation is quite different: here a direct encoding of the *d-ccar* condition leads to a rather natural class of procedural models. The class of models described next could be called, in analogy to Example 4.3, "direct *ccar* models". Since the models here described permit a more natural interpretation, we give it a different name, however.

**Example 4.4** *(Multiple grouped data model, MGD) Let $X$ be a $W$-valued random variable. Let $(\mathcal{W}_1, \ldots, \mathcal{W}_k)$ be a family of partitions of $W$ (cf. Theorem 2.14). Let $\boldsymbol{G} = G_1$, where $G_1$ takes values in $\{1, \ldots, k\}$ and is independent of $X$. Define $f(x, i)$ as that $U \in \mathcal{W}_i$ that contains $x$. Then $(X, G_1, f)$ is* ccar. *Conversely, every d-*ccar *coarse data model is induced by such a multiple grouped data model.*

The multiple grouped data model corresponds exactly to the CARGEN procedure of *GH*. It allows intuitive interpretations as representing procedures where one randomly selects one out of $k$ different available sensors or tests, each of which will reveal the true value of $X$ only up to the accuracy represented by the set $U \in \mathcal{W}_i$ containing $x$. In the special case $k = 1$ this corresponds to *grouped* or *censored* data (Heitjan & Rubin, 1991). *GH* introduced CARGEN as a procedure that is guaranteed to produce *d-car* distributions. They do not consider *d-ccar*, and therefore do not establish the exact correspondence between CARGEN and *d-ccar*. In a similar vein, *GvLR* introduced a general procedure for generating *d-car* distributions. The following example rephrases the construction of *GvLR* in our terminology.

**Example 4.5** *(Randomized monotone coarsening, RMC) Let $X$ be a $W$-valued random variable. Let $\boldsymbol{G} = H_1, S_1, H_2, S_2, \ldots, S_{n-1}, H_n$, where the $H_i$ take values in $2^W$, and the $S_i$ are $\{0, 1\}$-valued. Define*

$$\bar{H}_i := \begin{cases} H_i & \text{if } X \in H_i \\ W \setminus H_i & \text{if } X \notin H_i. \end{cases}$$

*Let the conditional distribution of $H_i$ given $X$ and $H_1, \ldots, H_{i-1}$ be concentrated on subsets of $\cap_{j=1}^{i-1} \bar{H}_j$.*

*This model represents a procedure where one successively refines a "current" coarse data set $\mathcal{A}_i := \cap_{h=1}^{i-1} \bar{H}_h$ by selecting a random subset $H_i$ of $\mathcal{A}_i$ and checking whether $X \in H_i$ or not, thus computing $\bar{H}_i$ and $\mathcal{A}_{i+1}$. This process is continued until for the first time $S_i = 1$ (i.e. the $S_i$ represent stopping conditions). The result of the procedure, then is represented by the following function $f$:*

$$f(X, (H_1, S_1, \ldots, S_{n-1}, H_n)) = \cap_{i=1}^{\min\{k | S_k = 1\}} \bar{H}_i.$$

*Finally, we impose the conditional independence condition that the distribution of the $H_i, S_i$ depend on $X$ only through $\bar{H}_1, \ldots, \bar{H}_{i-1}$, i.e.*

$$P(H_i \mid X, H_1, \ldots, H_{i-1}) = P(H_i \mid \bar{H}_1, \ldots, \bar{H}_{i-1})$$
$$P(S_i \mid X, H_1, \ldots, H_{i-1}) = P(S_i \mid \bar{H}_1, \ldots, \bar{H}_{i-1}).$$

As shown in *GvLR*, an *RMC* model always generates a *d-car* distribution, but not every *d-car* distribution can be obtained in this way. *GH* state that *RMC* models are a special case of CARGEN models. As we will see below, CARGEN and *RMC* are actually equivalent, and thus, both correspond exactly to *d-ccar* distributions. The distribution of Example 2.17 is the standard example (already used in a slightly different form in *GvLR*) of a *d-car* distribution that cannot be generated by *RMC* or CARGEN. A question of considerable interest, then, is whether there exist natural procedural models that correspond exactly to *d-car* distributions. *GvLR* state that they "cannot conceive of a more general mechanism than a randomized monotone coarsening scheme for constructing the *car* mechanisms which one would expect to meet with in practice,..." (p.267). *GH*, on the other hand, generalize the CARGEN models to a class of models termed CARGEN*, and show that these exactly comprise the models inducing *d-car* distributions.

However, the exact extent to which CARGEN* is more natural or reasonable than the trivial direct *car* models has not been formally characterized. We will discuss this issue below. First we present another class of procedural models. This is a rather intuitive class which contains models not equivalent to any CARGEN/*RMC* model.

**Example 4.6** (Uniform noise model) *Let $X$ be a $W$-valued random variable. Let $\boldsymbol{G} = N_1, H_1, N_2, H_2, \ldots$, where the $N_i$ are $\{0, 1\}$-valued, and the $H_i$ are $W$-valued with*

$$P(H_i = x) = 1/\left|W\right| \quad (x \in W). \tag{27}$$

*Let $X, N_1, H_1, \ldots$ be independent. Define for $h_i \in W, n_i \in \{0, 1\}$:*

$$f(x, (n_1, h_1, \ldots)) = \{x\} \cup \{h_i \mid i : n_i = 1\}. \tag{28}$$

*This model describes a procedure where in several steps (perhaps infinitely many) uniformly selected states from $W$ are added as noise to the observation. The random variables $N_i$ represent events that cause additional noise to be added. The distributions generated by this procedure are $d$-car, because for all $x, U$ with $x \in U$:*

$$P(Y = U \mid X = x) = P(\{h_i \mid i : n_i = 1\} = U) + P(\{h_i \mid i : n_i = 1\} = U \setminus \{x\}).$$

*By the uniformity condition (27), and the independence of the family $\{X, N_1, H_1, \ldots\}$, the last probability term in this equation is constant for $x \in U$.*

*The uniform noise model can not generate exactly the $d$-car distribution of Example 2.17. However, it can generate the variant of that distribution that was originally given in GvLR.*

The uniform noise model is rather specialized, and far from being able to induce every possible *d-car* distribution. As mentioned above, *GH* have proposed a procedure called CARGEN* for generating exactly all *d-car* distributions. This procedure is described in *GH* in the form of a randomized algorithm, but it can easily be recast in the form of a procedural model in the sense of Definition 4.1. We shall not pursue this in detail, however, and instead present a procedure that has the same essential properties as CARGEN* (especially with regard to the formal "reasonableness conditions" we shall introduce below), but is somewhat simpler and perhaps slightly more intuitive.

**Example 4.7** *(Propose and test model, P&T))* *Let $X$ be a $W$-valued random variable. Let $\boldsymbol{G} = G_1, G_2, \ldots$ be an infinite sequence of random variables taking values in $2^W \setminus \emptyset$. Let $X, G_1, G_2, \ldots$ be independent, and the $G_i$ be identically distributed, such that*

$$\sum_{U : x \in U} P(G_i = U) \quad \text{is constant on} \quad \{x \in W \mid P(X = x) > 0\}. \tag{29}$$

*Define*

$$f(x, (U_1, U_2, \ldots)) := \left\{ \begin{array}{ll} U_i & \text{if } i = \min\{j \geq 1 \mid x \in U_j\} \\ W & \text{if } \{j \geq 1 \mid x \in U_j\} = \emptyset \end{array} \right. .$$

The P&T model describes a procedure where we randomly propose a set $U \subseteq W$, test whether $x \in U$, and return $U$ if the result is positive (else continue). The condition (29) can be understood as an unbiasedness condition, which ensures that for every $x \in W$ (with $P(X = x) > 0$) we are equally likely to draw a positive test for $x$. The following theorem is analogous to Theorem 4.9 in *GH*; the proof is much simpler, however.

**Theorem 4.8** *A coarse data distribution $P$ is d-car iff it can be induced by a P&T model.*

**Proof:** That every distribution induced by a P&T model is *d-car* follows immediately from

$$P(Y = U \mid X = x) = P(G_i = U) / \sum_{U' : x \in U'} P(G_i = U'). \tag{30}$$

By (29) this is constant on $\{x \in U \mid P(X = x) > 0\}$ (note, too, that (29) ensures that the sum in the denominator of (30) is nonzero for all $x$, and that in the definition of $f$ the case $\{j \geq 1 \mid x \in U_j\} = \emptyset$ only occurs with probability zero).

Conversely, let $P$ be a *d-car* distribution on $\Omega(W)$. Define $c := \sum_{U \in 2^W} P(Y = U \mid X \in U)$, and

$$P(G_i = U) = P(Y = U \mid X \in U)/c.$$

Since $P(Y = U \mid X \in U) = P(Y = U \mid X = x)$ for all $x \in U$ with $P(X = x) > 0$, we have $\sum_{U : x \in U} P(Y = U \mid X \in U) = 1$ for all $x \in W$ with $P(X = x) > 0$. It follows that (29) is satisfied with $1/c$ being the constant. The resulting P&T model induces the original $P$:

$$P(f(X, \boldsymbol{G}) = U \mid X = x) = (P(Y = U \mid X \in U)/c) / (\sum_{U' : x \in U'} P(Y = U' \mid X \in U')/c)$$

$$= P(Y = U \mid X \in U) = P(Y = U \mid X = x).$$

$\square$

The P&T model looks like a reasonable natural procedure. However, it violates a desideratum that *GvLR* have put forward for a natural coarsening procedure:

> *(D)* In the coarsening procedure, no more information about the true value of $X$ should be used than is finally revealed by the coarse data variable $Y$ (Gill et al., 1997, p.266, paraphrased).

The P&T model violates desideratum *(D)*, because when we first unsuccessfully test $U_1, \ldots,$ $U_k$, then we require the information $x \notin \cup_{i=1}^{k} U_i$, which is not included in the final data $Y = U_{k+1}$. The observation generating process of Example 2.17, too, appears to violate *(D)*, as the host requires the precise value of $X$ when following his strategy. Finally, the uniform noise model violates *(D)*, because in the computation (28) of the final coarse data output the exact value of $X$ is required. These examples suggest that *(D)* is not a condition that one must necessarily expect every natural coarsening procedure to possess. *(D)* is most appropriate when coarse data is generated by an experimental process that is aimed at determining the true value of $X$, but may be unable to do so precisely. In such a scenario, *(D)* corresponds to the assumption that all information about the value of $X$ that is collected in the experimental process also is reported in the final result. Apart from experimental procedures, also 'accidental' processes corrupting complete data can generate *d-car* data (as represented, e.g., by the uniform noise model). For such procedures *(D)* is not immediately seen as a necessary feature. However, Theorem 4.17 below will lend additional support to *(D)* also in these cases.

*GH* argue that their class of CARGEN* procedures only contains reasonable processes, because "each step of the algorithm can depend only on information available to the experimenter, where the 'information' is encoded in the observations made by the experimenter in the course of running the algorithm"(*GH*, p. 260). The same can be said about the P&T procedure. The direct *car* model would not be reasonable in this sense, because for the simulation of the variable $G$ one would need to pick a distribution dependent on the true value of $X$, which is not assumed to be available. However, it is hard to make rigorous this distinction between direct *car* models on the one hand, and CARGEN*/P&T on the other hand, because the latter procedures permit tests for the value of $X$ (through checking $X \in U$ for test sets $U$ – using singleton sets $U$ one can even query the exact value of $X$), and the continuation of the simulation is dependent on the outcome of these tests.

We will now establish a more solid foundation for discussing reasonable vs. unreasonable coarsening procedures by introducing two different rigorous conditions for natural or reasonable *car* procedures. One is a formalization of desideratum *(D)*, while the other expresses an invariance of the *car* property under numerical parameter changes. We will then show that these conditions can only be satisfied when the generated distribution is *d-ccar*. For the purpose of this analysis it is helpful to restrict attention to a special type of procedural models.

**Definition 4.9** *A procedural model* $(X, \boldsymbol{G}, f)$ *is a* Bernoulli-model *if the family* $X, G_1,$ $G_2, \ldots$ *is independent.*

The name Bernoulli model is not quite appropriate here, because the variables $X, G_i$ are not necessarily binary. However, it is clear that one could also replace the multinomial $X$ and $G_i$ with suitable sets of (independent) binary random variables. In essence, then, a Bernoulli model in the sense of Definition 4.9 can be seen as an infinite sequence of independent coin tosses (with coins of varying bias). Focusing on Bernoulli models is no real limitation:

**Theorem 4.10** *Let* $(X, \boldsymbol{G}, f)$ *be a procedural model. Then there exists a Bernoulli model* $(X, \boldsymbol{G}^*, f^*)$ *inducing the same coarse data distribution.*

The reader may notice that the statement of Theorem 4.10 really is quite trivial: the coarse data distribution induced by $(X, \boldsymbol{G}, f)$ is just a distribution on the finite coarse data space $\Omega(W)$, and there are many simple, direct constructions of Bernoulli models for such a given distribution. The significance of Theorem 4.10, therefore, lies essentially in the following proof, where we construct a Bernoulli model $(X, \boldsymbol{G}^*, f^*)$ that preserves all the essential procedural characteristics of the original model $(X, \boldsymbol{G}, f)$. In fact, the model $(X, \boldsymbol{G}^*, f^*)$ can be understood as an implementation of the procedure $(X, \boldsymbol{G}, f)$ using a generator for independent random numbers.

To understand the intuition of the construction, consider a randomized algorithm for simulating the procedural model $(X, \boldsymbol{G}, f)$. The algorithm successively samples values for $X, G_1, G_2, \ldots$, and finally computes $f$ (for most natural procedural models the value of $f$ is already determined by finitely many initial $G_i$-values, so that not infinitely many $G_i$ need be sampled, and the algorithm actually terminates; for our considerations, however, algorithms taking infinite time pose no conceptual difficulties). The distribution used for sampling $G_i$ may depend on the values of previously sampled $G_1, \ldots, G_{i-1}$, which, in a computer implementation of the algorithm are encoded in the current program state.

The set of all possible runs of the algorithm can be represented as a tree, where branching nodes correspond to sampling steps for the $G_i$. A single execution of the algorithm generates one branch in this tree. One can now construct an equivalent algorithm that, instead, generates the whole tree breadth-first, and that labels each branching node with a random value for the $G_i$ associated with the node, sampled according to the distribution determined by the program state corresponding to that node. In this algorithm, sampling of random values is independent. The labeling of all branching nodes identifies a unique branch in the tree, and for each branch, the probability of being identified by the labeling is equal to the probability of this branch representing the execution of the original algorithm (a similar transformation by pre-computing all random choices that might become relevant is described in by Gill & J.M.Robins, 2001[Section 7]). The following proof formalizes the preceding informal description.

**Proof of Theorem 4.10:** For each random variable $G_i$ we introduce a sequence of random variables $G_{i,1}^*, \ldots, G_{i,K(i)}^*$, where $K(i) = |W \times \times_{j=1}^{i-1} \Gamma_j|$ is the size of the joint state space of $X, G_1, \ldots, G_{i-1}$. The state space of the $G_{i,h}^*$ is $\Gamma_i$ (with regard to our informal explanation, $G_{i,h}^*$ corresponds to the node in the full computation tree that represent the sampling of $G_i$ when the previous execution has resulted in the $h$th out of $K(i)$ possible program states). We construct a joint distribution for $X$ and the $G_{i,h}^*$ by setting $P(G_{i,h}^* = v) = P(G_i = v \mid (X, G_1, \ldots, G_{i-1}) = s_h)$ ($s_h$ the $h$th state in an enumeration of $W \times \times_{j=1}^{i-1} \Gamma_j$), and by taking $X$ and the $G_{i,h}^*$ to be independent.

It is straightforward to define a mapping

$$h^* : W \times \times_{i \geq 1} \Gamma_i^{K(i)} \to \boldsymbol{\Gamma}$$

such that $(X, h^*(X, \boldsymbol{G}^*))$ is distributed as $(X, \boldsymbol{G})$ (the mapping $h^*$ corresponds to the extraction of the "active" branch in the full labeled computation tree). Defining $f^*(x, \boldsymbol{g}^*) := f(x, h^*(x, \boldsymbol{g}^*))$ then completes the construction of the Bernoulli model. $\square$

**Definition 4.11** *The Bernoulli model* $(X, \boldsymbol{G}^*, f^*)$ *obtained via the construction of the proof of Theorem 4.10 is called the* Bernoulli transform *of* $(X, \boldsymbol{G}, f)$.

**Example 4.12** *For a direct* car *model* $(X, G, f)$ *we obtain the Bernoulli transform* $(X, (G_1^*, \ldots, G_n^*), f^*)$, *where*

$$P(G_i^* = U) = P(G = U \mid X = x_i),$$
$$h^*(x_i, U_1, \ldots, U_n) = (x_i, U_i),$$

*and so* $f^*(x_i, U_1, \ldots, U_n) = U_i$.

When the coarsening procedure is a Bernoulli model, then no information about $X$ is used for sampling the variables $\boldsymbol{G}$. The only part of the procedure where $X$ influences the outcome is in the final computation of $Y = f(X, \boldsymbol{G})$. The condition that in this computation only as much knowledge of $X$ should be required as finally revealed by $Y$ now is basically condition (9) for coarsening variables. The state space $\boldsymbol{\Gamma}$ for $\boldsymbol{G}$ now being (potentially) uncountable, it is however more appropriate to replace the universal quantification "for all $g$" in (9) with "for almost all $g$" in the probabilistic sense. We thus define:

**Definition 4.13** *A Bernoulli model is* honest, *if for all* $x, x'$ *with* $P(X = x) > 0, P(X = x') > 0$, *and all* $U \in 2^W \setminus \emptyset$:

$$P(\boldsymbol{G} \in \{\boldsymbol{g} \mid f(x, \boldsymbol{g}) = U, x' \in U \Rightarrow f(x', \boldsymbol{g}) = U\}) = 1. \tag{31}$$

**Example 4.14** *The Bernoulli model of Example 4.12 is not honest, because one can have for some* $U_1, \ldots, U_n$ *with* $P(\boldsymbol{G} = (U_1, \ldots, U_n)) > 0$: $U_j \neq U_i$, *and* $x_i, x_j \in U_i$, *such that*

$$f^*(x_i, U_1, \ldots, U_n) = U_i \neq U_j = f^*(x_j, U_1, \ldots, U_n).$$

Honest Bernoulli models certainly satisfy *(D)*. On the other hand, there can be non-Bernoulli models that also seem to satisfy *(D)* (notably the RMC models, which were developed with *(D)* in mind). However, for non-Bernoulli models it appears hard to make precise the condition that the sampling of $\boldsymbol{G}$ does not depend on $X$ beyond the fact that $X \in Y$ [1]. The following theorem indicates that our formalization of *(D)* in terms of Bernoulli models only is not too narrow.

**Theorem 4.15** *The Bernoulli transforms of MGD, CAR*GEN *and RMC models are honest.*

The proof for all three types of models are elementary, though partly tedious. We omit the details here.

We now turn to a second condition for reasonable procedures. For this we observe that the MGD/CAR GEN/RMC models are essentially defined in terms of the "mechanical procedure" for generating the coarse data, whereas the direct *car*, the uniform noise, and the P&T models (and in a similar way CAR GEN*) rely on the numerical conditions (26),(27), respectively (29), on distributional parameters. These procedures, therefore, are fragile in the sense that slight perturbations of the parameters will destroy the *d-car* property of the induced distribution. We would like to distinguish robust *car* procedures as those for which the *d-car* property is guaranteed through the mechanics of the process alone (as determined by the state spaces of the $G_i$, and the definition of $f$), and does not depend on parameter constraints for the $G_i$ (which, in a more or less subtle way, can be used to mimic the brute force condition (26)). Thus, we will essentially consider a *car* procedure to be robust, if it stays *car* under changes of the parameter settings for the $G_i$. There are two points to consider before we can state a formal definition of this idea. First, we observe that our concept of robustness should again be based on Bernoulli models, since in non-Bernoulli models even arbitrarily small parameter changes can create or destroy independence relations between the variables $X, \boldsymbol{G}$, and such independence relations, arguably, reflect qualitative rather than merely quantitative aspects of the coarsening mechanism.

---

1. The intuitive condition that $\boldsymbol{G}$ must be independent of $X$ given $Y$ turns out to be inadequate.

Secondly, we will want to limit permissible parameter changes to those that do not lead to such drastic quantitative changes that outcomes with previously nonzero probability become zero-probability events, or vice versa. This is in line with our perspective in Section 3, where the set of support of a distribution on a finite state space was viewed as a basic qualitative property. In our current context we are dealing with distributions on uncountable state spaces, and we need to replace the notion of identical support with the notion of absolute continuity: recall that two distributions $P, \tilde{P}$ on a state space $\Sigma$ are called *mutually absolutely continuous*, written $P \equiv \tilde{P}$, if $P(S) = 0 \Leftrightarrow \tilde{P}(S) = 0$ for all measurable $S \subseteq \Sigma$.

For a distribution $P(\boldsymbol{G})$ on $\boldsymbol{\Gamma}$, with $\boldsymbol{G}$ an independent family, we can obtain $\tilde{P}(\boldsymbol{G})$ with $P \equiv \tilde{P}$, for example, by changing for finitely many $i$ parameter values $P(G_i = g) = r > 0$ to new values $\tilde{P}(G_i = g) = \tilde{r} > 0$. On the other hand, if e.g. $\Gamma_i = \{0, 1\}$, $P(G_i = 0) = 1/2$, and $\tilde{P}(G_i = 0) = 1/2 + \epsilon$ for all $i$ and some $\epsilon > 0$, then $P(\boldsymbol{\Gamma}) \not\equiv \tilde{P}(\boldsymbol{\Gamma})$. For a distribution $P(X)$ of $X$ alone one has $P(X) \equiv \tilde{P}(X)$ iff $P$ and $\tilde{P}$ have the same support.

**Definition 4.16** *A Bernoulli model $(X, \boldsymbol{G}, f)$ is robust car (robust ccar), if it is car (ccar), and remains car (ccar) if the distributions $P(X)$ and $P(G_i)$ $(i \geq 1)$ are replaced with distributions $\tilde{P}(X)$ and $\tilde{P}(G_i)$, such that $P(X) \equiv \tilde{P}(X)$ and $P(\boldsymbol{G}) \equiv \tilde{P}(\boldsymbol{G})$.*

The Bernoulli transforms of MGD/CAR$_{\text{GEN}}$ are robust *ccar*. Of the class RMC we know, so far, that it is *car*. The Bernoulli transform of RMC can be seen to be robust *car*. The Bernoulli transforms of CAR$_{\text{GEN}}$*/P&T, on the other hand, are not robust (and neither is the uniform noise model, which already is Bernoulli). We now come to the main result of this section, which basically identifies the existence of 'reasonable' procedural models with *d-ccar*.

**Theorem 4.17** *The following are equivalent for a distribution $P$ on $\Omega(W)$:*

**(i)** *$P$ is induced by a robust car Bernoulli model.*

**(ii)** *$P$ is induced by a robust ccar Bernoulli model.*

**(iii)** *$P$ is induced by an honest Bernoulli model.*

**(iv)** *$P$ is d-ccar.*

The proof is given in Appendix A. Theorem 4.17 essentially identifies the existence of a natural procedural model for a *d-car* distribution with the property of being *d-ccar*, rather than merely *d-car*. This is a somewhat surprising result at first sight, given that $\boldsymbol{M}$*-mcar* is usually considered an unrealistically strong assumption as compared to $\boldsymbol{M}$*-mar*. There is no real contradiction here, however, as we have seen before that *d-ccar* is weaker than $\boldsymbol{M}$*-mcar*. Theorem 4.17 indicates that in practice one may find many cases where *d-ccar* holds, but $\boldsymbol{M}$*-mcar* is not fulfilled.

## 5. Conclusion

We have reviewed several versions of *car* conditions. They differ with respect to their formulation, which can be in terms of a coarsening variable, or in terms of a purely distributional constraint. The different versions are mostly non-equivalent. Some care, therefore, is required in determining for a particular statistical or probabilistic inference problem the appropriate *car* condition that is both sufficient to justify the intended form of inference, and the assumption of which is warranted for the observational process at hand. We argue that the distributional forms of *car* are the more relevant ones: when the observations are fully described as subsets of $W$, then the coarse data distribution is all that is required in the analysis, and the introduction of an artificial coarsening variable $G$ can skew the analysis.

Our main goal was to provide characterizations of coarse data distributions that satisfy *d-car*. We considered two types of such characterizations: the first type is a "static" description of *d-car* distributions in terms of their sets of support. Here we have derived a quite intuitive, complete characterization by means of the support hypergraph of a coarse data distribution.

The second type of characterizations is in terms of procedural models for the observational process that generates the coarse data. We have considered several models for such observational processes, and found that the arguably most natural ones are exactly those that generate observations which are *d-ccar*, rather than only *d-car*. This is somewhat surprising at first, because $\boldsymbol{M}$*-ccar* is typically an unrealistically strong assumption (cf. Example 2.3). The distributional form, *d-ccar*, on the contrary, turns out to be the perhaps most natural assumption. The strongest support support for the *d-ccar* assumption is provided by the equivalence $(i) \Leftrightarrow (iv)$ in Theorem 4.17: assuming *d-car*, but not *d-ccar*, means that we must be dealing with a fragile coarsening mechanism that produces *d-car* data only by virtue of some specific parameter settings. Since we usually do not know very much about the coarsening mechanism, the assumption of such a special parameter-equilibrium (as exemplified by (29)) will typically be unwarranted.

## Acknowledgments

The author would like to thank Ian Pratt for providing the initial motivation for investigating the basis of probabilistic inference by conditioning. Richard Gill, Peter Grünwald, and James Robins have provided valuable comments to earlier versions of this paper. I am particularly indebted to Peter Grünwald for suggestions on the organization of the material in Section 2.1, which led to a great improvement in the presentation. Richard Gill must be credited for the short proof of Theorem 3.10, which replaced a previous much more laborious one.

## Appendix A. Proof of Theorem 4.17

**Theorem 4.17** The following are equivalent for a distribution $P$ on $\Omega(W)$:

**(i)** $P$ is induced by a robust *car* Bernoulli model.

**(ii)** $P$ is induced by a robust *ccar* Bernoulli model.

**(iii)** $P$ is induced by an honest Bernoulli model.

**(iv)** $P$ is *d-ccar*.

We begin with some measure theoretic preliminaries. Let $\mathcal{A}$ be the product $\sigma$-algebra on $\boldsymbol{\Gamma}$ generated by the powersets $2^{\Gamma_i}$. The joint distribution $P(X, \boldsymbol{G})$ then is defined on the product of $2^W$ and $\mathcal{A}$. The $\sigma$-algebra $\mathcal{A}$ is generated by the *cylinder sets* $(g_1^*, g_2^*, \ldots, g_k^*) \times \times_{j>k} \Gamma_j$ ($k \geq 0$, $g_h^* \in \Gamma_h$ for $h = 1, \ldots, k$). The cylinder sets also are the basis for a topology $\mathcal{O}$ on $\boldsymbol{\Gamma}$. The space $(\boldsymbol{\Gamma}, \mathcal{O})$ is compact (this can be seen directly, or by an application of Tikhonov's theorem). It follows that every probability distribution $P$ on $\mathcal{A}$ is *regular*, especially for all $A \in \mathcal{A}$:

$$P(A) = \inf\{P(O) \mid A \subseteq O \in \mathcal{O}\}$$

(see e.g. Cohn, 1993, Prop. 7.2.3). Here and in the following we use interchangeably the notation $P(A)$ and $P(\boldsymbol{G} \in A)$. The former notation is sufficient for reasoning about probability distributions on $\mathcal{A}$, the latter emphasizes the fact that we are always dealing with distributions induced by the family $\boldsymbol{G}$ of random variables.

**Lemma A.1** *Let $P(\boldsymbol{G})$ be the joint distribution on $\mathcal{A}$ of an independent family $\boldsymbol{G}$. Let $A_1, A_2 \in \mathcal{A}$ with $A_1 \cap A_2 = \emptyset$ and $P(\boldsymbol{G} \in A_1) = P(\boldsymbol{G} \in A_2) > 0$. Then there exists a joint distribution $\tilde{P}(\boldsymbol{G})$ with $P(\boldsymbol{G}) \equiv \tilde{P}(\boldsymbol{G})$ and $\tilde{P}(\boldsymbol{G} \in A_1) \neq \tilde{P}(\boldsymbol{G} \in A_2)$.*

**Proof:** Let $p := P(A_1)$. Let $\epsilon = p/2$ and $O \in \mathcal{O}$ such that $A_1 \subseteq O$ and $P(O) < p + \epsilon$. Using the disjointness of $A_1$ and $A_2$ one obtains $P(A_1 \mid O) > P(A_2 \mid O)$. Since the cylinder sets are a basis for $\mathcal{O}$, we have $O = \cup_{i \geq 0} Z_i$ for a countable family of cylinders $Z_i$. It follows that also for some cylinder set $Z = (g_1^*, g_2^*, \ldots, g_k^*) \times \times_{j>k} \Gamma_j$ with $P(Z) > 0$: $P(A_1 \mid Z) > P(A_2 \mid Z)$. Now let $\delta > 0$ and define for $h = 1, \ldots, k$:

$$\tilde{P}(G_h = g_h^*) := 1 - \delta; \quad \tilde{P}(G_h = g) := \delta(P(G_h = g) / \sum_{g' : g' \neq g_h^*} P(G_h = g')) \ (g \neq g_h^*)$$

For $h \geq k+1$: $\tilde{P}(G_h) := P(G_h)$. Then $P(\boldsymbol{G}) \equiv \tilde{P}(\boldsymbol{G})$, $\tilde{P}(A_1 \mid Z) = P(A_1 \mid Z)$, $\tilde{P}(A_2 \mid Z) = P(A_2 \mid Z)$, and therefore:

$$\tilde{P}(A_1) \geq (1 - \delta)^k P(A_1 \mid Z), \quad \tilde{P}(A_2) \leq (1 - \delta)^k P(A_2 \mid Z) + 1 - (1 - \delta)^k.$$

For sufficiently small $\delta$ this gives $\tilde{P}(A_1) > \tilde{P}(A_2)$. $\square$

**Proof of Theorem 4.17:** For simplification we may assume that $P(x) > 0$ for all $x \in W$. This is justified by the observation that none of the conditions (i)-(iv) are affected by adding or deleting states with zero probability from $W$.

The implication (iv)⇒(ii) follows from Example 4.4 by the observation that MGD models are robust *d-ccar* Bernoulli models. (ii)⇒(i) is trivial. We will show (i)⇒(iii) and (iii)⇒(iv).

First assume (i), and let $(X, \boldsymbol{G}, f)$ be a robust *car* Bernoulli model inducing $P$. For $x \in W$ and $U \subseteq W$ denote

$$A(x, U) := \{\boldsymbol{g} \in \boldsymbol{\Gamma} \mid f(x, \boldsymbol{g}) = U\}.$$

The *d-car* property of $P$ is equivalent to

$$P(\boldsymbol{G} \in A(x, U)) = P(\boldsymbol{G} \in A(x', U)). \tag{32}$$

for all $x, x' \in U$.

Condition (31) is equivalent to the condition that $P(\boldsymbol{G} \in A(x, U) \setminus A(x', U)) = 0$ for $x, x' \in U$. Assume otherwise. Then for $A_1 := A(x, U) \setminus A(x', U)$, $A_2 := A(x', U) \setminus A(x, U)$: $0 < P(\boldsymbol{G} \in A_1) = P(\boldsymbol{G} \in A_2)$. Applying Lemma A.1 we obtain a Bernoulli model $\tilde{P}(X, \boldsymbol{G}) = P(X)\tilde{P}(\boldsymbol{G})$ with $\tilde{P}(X, \boldsymbol{G}) \equiv P(X, \boldsymbol{G})$ and $\tilde{P}(\boldsymbol{G} \in A_1) \neq \tilde{P}(\boldsymbol{G} \in A_2)$. Then also $\tilde{P}(\boldsymbol{G} \in A(x, U)) \neq \tilde{P}(\boldsymbol{G} \in A(x', U))$, so that $\tilde{P}(X, \boldsymbol{G})$ is not *d-car*, contradicting (i).

(iii)⇒(iv): Let

$$\boldsymbol{\Gamma}^* := \bigcap_{\substack{x, x' \in W, U \subseteq W: \\ x, x' \in U}} \{\boldsymbol{g} \mid f(x, \boldsymbol{g}) = U, x' \in U \Rightarrow f(x', \boldsymbol{g}) = U\}.$$

Since the intersection is only over finitely many $x, x', U$, we obtain from (iii) that $P(\boldsymbol{G} \in \boldsymbol{\Gamma}^*) = 1$. For $U \subseteq W$ define $A(U) := A(x, U) \cap \boldsymbol{\Gamma}^*$, where $x \in U$ is arbitrary. By the definition of $\boldsymbol{\Gamma}^*$ the definition of $A(U)$ is independent of the particular choice of $x$. Define an equivalence relation $\sim$ on $\boldsymbol{\Gamma}^*$ via

$$\boldsymbol{g} \sim \boldsymbol{g}' \quad \Leftrightarrow \quad \forall U \subseteq W: \; \boldsymbol{g} \in A(U) \Leftrightarrow \boldsymbol{g}' \in A(U). \tag{33}$$

This equivalence relation partitions $\boldsymbol{\Gamma}^*$ into finitely many equivalence classes $\boldsymbol{\Gamma}_1^*, \ldots, \boldsymbol{\Gamma}_k^*$. We show that for each $\boldsymbol{\Gamma}_i^*$ and $\boldsymbol{g} \in \boldsymbol{\Gamma}_i^*$ the system

$$\mathcal{W}_i := \{U \mid \exists x \in W: \; f(x, \boldsymbol{g}) = U\} \tag{34}$$

is a partition of $W$, and that the definition of $\mathcal{W}_i$ does not depend on the choice of $\boldsymbol{g}$. The latter claim is immediate from the fact that for $\boldsymbol{g} \in \boldsymbol{\Gamma}^*$

$$f(x, \boldsymbol{g}) = U \quad \Leftrightarrow \quad \boldsymbol{g} \in A(U) \text{ and } x \in U. \tag{35}$$

For the first claim assume that $f(x, \boldsymbol{g}) = U, f(x', \boldsymbol{g}) = U'$ with $U \neq U'$. In particular, $\boldsymbol{g} \in A(U) \cap A(U')$. Assume there exists $x'' \in U \cap U'$. Then by (35) we would obtain both $f(x'', \boldsymbol{g}) = U$ and $f(x'', \boldsymbol{g}) = U'$, a contradiction. Hence, the sets $U$ in the $\mathcal{W}_i$ are pairwise disjoint. They also are a cover of $W$, because for every $x \in W$ there exists $U$ with $x \in U = f(x, \boldsymbol{g})$.

We thus obtain that the given Bernoulli model is equivalent to the multiple grouped data model defined by the partitions $\mathcal{W}_i$ and parameters $\lambda_i := P(\boldsymbol{G} \in \boldsymbol{\Gamma}_i^*)$. □